\documentclass[10pt,letterpaper]{article}
\usepackage[utf8]{inputenc} 
\usepackage[T1]{fontenc}    
\usepackage{hyperref}       
\usepackage{url}            
\usepackage{booktabs}       
\usepackage{amsfonts}       
\usepackage{nicefrac}       
\usepackage{microtype}      
\usepackage{graphicx}
\usepackage{amssymb,amsmath}
\usepackage{mathtools}
\usepackage{cogsci}

\usepackage{todonotes}
\usepackage{pslatex}
\usepackage{apacite}
\usepackage{float}
\usepackage{stmaryrd}

\graphicspath{ {./figures/} }

\title{Learning to communicate about shared procedural abstractions}

\cogscifinaltrue
\author{
  \large \bf William P. McCarthy\textsuperscript{*}\\
  Department of Cognitive Science\\
  UC San Diego\\
  \texttt{wmccarthy@ucsd.edu} \\
  \And
  \large \bf Robert D. Hawkins\textsuperscript{*}\\
  Department of Psychology\\
  Princeton University\\
  \texttt{rdhawkins@princeton.edu} \\
  \And
  \large \bf Haoliang Wang\\
  Department of Psychology\\
  UC San Diego\\
  \texttt{haw027@ucsd.edu} \\
  \AND
  \large \bf Cameron Holdaway\\
  Department of Psychology\\
  UC San Diego\\
  \texttt{choldawa@ucsd.edu} \\
  \And
  \large \bf Judith E. Fan\\
  Department of Psychology\\
  UC San Diego\\
  \texttt{jefan@ucsd.edu} \\
}

\begin{document}

\maketitle

\begin{abstract}

Many real-world tasks require agents to coordinate their behavior to achieve shared goals.
Successful collaboration requires not only adopting the same communicative conventions, but also grounding these conventions in the same task-appropriate conceptual abstractions. 
We investigate how humans use natural language to collaboratively solve physical assembly problems more effectively over time.
Human participants were paired up in an online environment to reconstruct scenes containing two block towers. 
One participant could see the target towers, and sent assembly instructions for the other participant to reconstruct.
Participants provided increasingly concise instructions across repeated attempts on each pair of towers, using more abstract referring expressions that captured each scene's hierarchical structure.
To explain these findings, we extend recent probabilistic models of \emph{ad hoc} convention formation with an explicit perceptual learning mechanism. 
These results shed light on the inductive biases that enable intelligent agents to coordinate upon shared procedural abstractions.

\end{abstract}
\let\thefootnote\relax\footnote{* denotes equal contribution}

From advanced manufacturing to food preparation, many real-world tasks require multiple agents to coordinate their behavior \cite{grosz1996collaborative,stone2010ad,wang2020too}. 
To coordinate effectively, collaborators benefit from sharing similar representations of relevant objects and procedures, specified at the appropriate level of abstraction for their joint goals.
For example, when a new cook is training in a kitchen, they may need to follow step-by-step instructions at the level of individual ingredients, like \emph{melt 30g butter in the small pan, then stir in 30g of flour}.
As they gain more experience, however, they may just \emph{make a roux}, efficiently executing the entire procedure as a single routine. 
When all cooks are using the same unified \emph{roux} abstraction, this simplifies coordination in the kitchen in several ways.
First, they are able to plan more efficiently when they expect other agents to follow chunked routines, since it is no longer necessary to consider all possible low-level executions.
Second, they no longer need to divide up sub-tasks (e.g. one agent melting the butter and the other agent stirring in the flour) when agents can be mutually expected to follow a unitized sub-routine to completion.

In many cases, however, these abstractions are not supplied to agents in advance, and achieving their collective benefits requires \textit{ad hoc} coordination between agents as they each learn about what is required for the task \cite{wang2017naturalizing}.
A powerful solution to the problem of coordinating abstractions is the ability to communicate using natural language \cite{suhr2019executing,tellex2020robots}.
Yet for communication protocols to be effective in novel task settings, where there may not yet be words to easily express the task-specific abstractions, these protocols must \emph{also} be able to update over the course of an interaction, a phenomenon that has been explored in both psycholinguistics \cite{clark1996using, hawkins2020characterizing} and natural language processing \cite{hawkins2019continual}. 
How, then, are intelligent, autonomous agents able to simultaneously coordinate on shared object representations \emph{and} the language for talking about them?

\begin{figure*}[ht]
\begin{center}
\includegraphics[width=0.99\linewidth]{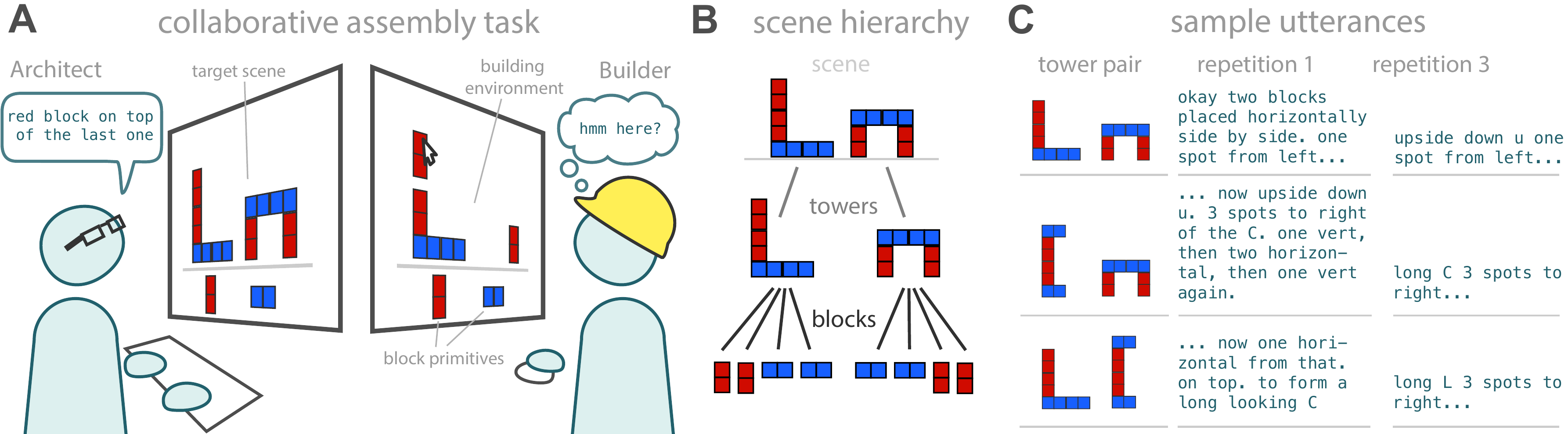}
\vspace{-1em}
\caption{Collaborative assembly task. (A) The Architect was shown a target scene and provided assembly instructions to the Builder, who aimed to reconstruct it. (B) Each scene was composed of two towers, which were each composed of four domino-shaped blocks. (C) Example messages from earlier and later repetitions of a tower pair, showing the emergence of expressions referring to towers.}
\vspace{-1em}
\label{fig:task}
\end{center}
\end{figure*}

In this paper, we approach this question by integrating two distinct computational approaches into a unified model.
On one hand, we draw on probabilistic models of \emph{ad hoc} lexical convention formation \cite{hawkinsgeneralizing} through interaction.
These models provide an account of how agents are able to coordinate on ways of referring to \emph{existing} conceptual primitives, but do not explain where new conceptual primitives come from.
On the other hand, recent models of perceptual learning as program synthesis \cite{gulwani2017program} have provided a powerful account of human conceptual representations.
These models propose that concepts may be represented by structured \emph{programs} written in a domain-specific language (DSL). 
Agents are able to supplement their library of primitive concepts with new abstractions, or chunked sub-routines, as they learn more about a task \cite{ellis2020dreamcoder}.
Importantly, these abstractions are \emph{compositional}, allowing them to be combined into larger programs with other primitives.

We suggest that these structured library learning mechanisms may supply agents with the raw conceptual primitives that ground new \emph{ad hoc} conventions in new tasks, and conversely, that communication may be an important mechanism that allows agents to coordinate their abstractions.
Here we explore this hypothesis by examining how humans coordinate their behavior in a physical assembly domain \cite{bapst2019structured, mccarthy2020blocks} in which objects are hierarchically organized, and can thus be specified at different levels of abstraction.
Overall, our paper presents an empirical paradigm, human dataset, and set of evaluation metrics that can be used to guide ongoing development of artificial agents that emulate human-like compositionality and abstraction.

\section{Collaborative assembly task}
\paragraph{Design, stimuli, and procedure}
We recruited 98 human participants ($N=49$ dyads) from Amazon Mechanical Turk and automatically paired them up to perform a collaborative assembly task (Fig.~\ref{fig:task}A).
At the outset, each participant was assigned the role of \textit{Architect} or \textit{Builder} and proceeded with their partner through a series of twelve trials.
At the start of each trial, the Architect was presented with a target scene containing block towers. 
The Builder could not see the target scene, and was presented with an empty grid world environment in which they could place blocks.
The Architect then sent step-by-step assembly instructions, which the Builder used to reconstruct the target scene as accurately as possible. 

Each scene was composed hierarchically from two block towers that appeared side by side; in turn, each tower consisted of four domino-shaped blocks-- two vertical and two horizontal (Fig.~\ref{fig:task}B).
To evaluate changes in behavior, we employed a \emph{repeated} design where each tower appeared multiple times. 
There were three unique towers. 
All three pairs of these towers appeared once in each of four repetition blocks in a randomized sequence, for a total of twelve trials. 
All towers appeared in both the left and right positions an equal number of times, such that there was no statistical association between a given tower and its position.

The Architect and Builder took as many turns as they needed to reconstruct each scene.
On the Architect's turn, they sent a single message containing a maximum of 100 characters; 
on the Builder's turn, they placed one or more blocks before awaiting further instructions (Fig.~\ref{fig:task}C). 
Blocks could be placed anywhere so long as they were supported from beneath, and could not be moved once placed.
The Architect could see the placement of each block in real time but the communication channel was otherwise unidirectional: the Builder was unable to send messages back to the Architect.
Once all eight blocks had been placed, both participants received feedback about the mismatch between the target scene and reconstruction before advancing to the next trial.

\begin{figure}[tbp]
\vspace{-1.5em}
\begin{center}
\includegraphics[width=0.8\linewidth]{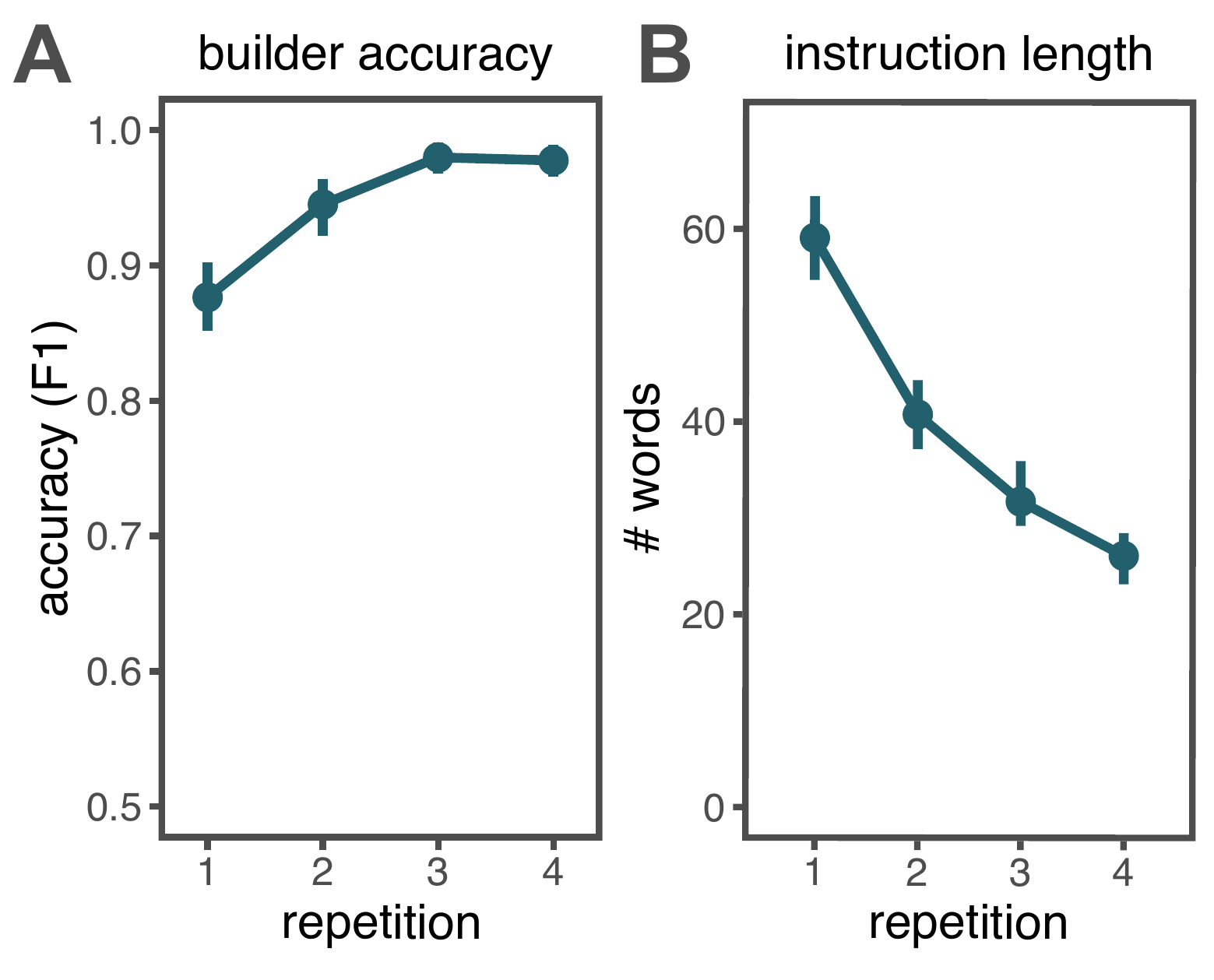}
\vspace{-1.5em}
\caption{(A) Reconstruction accuracy improved across repetitions. (B) Mean number of words used on each trial decreased across repetitions.}
\label{fig:performance}
\vspace{-1em}
\end{center}
\end{figure}

\begin{figure*}[tp]
\begin{center}
\includegraphics[width=\linewidth]{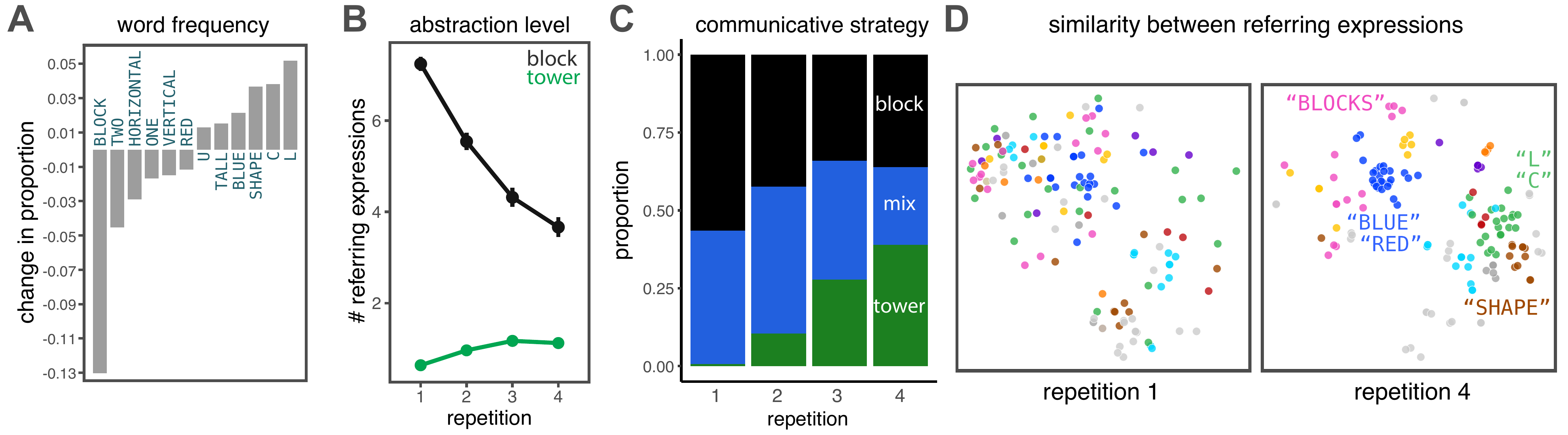}
\vspace{-2em}
\caption{(A) Words with largest positive and negative changes in frequency between first and final repetitions. (B) Change in number of block-level and tower-level references across repetitions. (C) The proportion of referring expressions in each trial that exclusively refer to blocks, towers, or scenes. (D) t-SNE visualization of similarity between messages from different dyads in the first and final repetitions.}
\label{fig:refexp}
\end{center}
\end{figure*}


\section{Behavioral Results}
Although each interaction only spanned twelve trials, we hypothesized that human dyads would be able to leverage this small amount of experience to rapidly develop shared task representations, manifesting in increasingly successful and efficient collaboration over time. 
\paragraph{Success across repetitions}
Given that the focus of our study was on how language produced by Architects changed over time, we sought to first verify that human dyads were able to successfully perform the assembly task. 
We found that even on their initial reconstructions, they were highly accurate (mean $F_1=0.876$; 95\%~CI:$[0.854, 0.898]$), which roughly corresponds to having just one block out of place.
Even so, we found that dyads reliably improved across repetitions ($b=3.38$, $t=7.90$, $p<0.001$; Fig.~\ref{fig:performance}A), estimated using a linear mixed-effects model that predicted accuracy from repetition number and included random intercepts for each dyad.  







\paragraph{Communicative efficiency across repetitions}
Given that the same towers recurred throughout the interaction, we hypothesized that Architects would exploit these regularities to provide more concise instructions over time. 
To test this hypothesis, we analyzed both changes in the total number of words used and how many messages were sent within a trial.
We estimated changes using LME models containing repetition number as a predictor, as well as random intercepts and slopes for each dyad and random intercepts for each tower pair.
Consistent with our hypothesis, we found that Architects sent messages containing fewer words over time ($b=-10.8$, $t=-10.9$, $p<0.001$) (Fig.~\ref{fig:performance}B), which were themselves contained in fewer messages within each trial ($b=-0.67$, $t=-8.01$, $p<0.001$).

\paragraph{Changes in words used across repetitions}
What explains these gains in communicative efficiency?
One possibility is that Architects increasingly omitted unnecessary, non-referential function words; another is that they changed which words they used to refer to objects. 
To distinguish these possibilities, we compared changes in the frequency of words used in the first and final repetitions. 
To ensure that our analyses reflected changes in the referring expression used to refer to components of each scene rather than in the use of function words, we recruited two human annotators who were blind to the source of each utterance to manually extract referring expressions from each message\footnote{Two dyads were excluded from this analysis because our annotators were unable to recover referring expressions from their language.}. 
For each dyad, we compared the word frequency distributions between the first and final repetitions using a permutation-based $\chi^2$ test \cite{beh2014correspondence}, which revealed a reliable difference between the two distributions ($p<0.001$, Bonferroni corrected for multiple comparisons).
To identify the words contributing most to this shift, we calculated the overall change in proportion from the first repetition to the final repetition.
We found that words such as ``block'' and ``horizontal'' were used less often while ``shape,'' and ``C'' were used more often (Fig.~\ref{fig:refexp}A).
These results suggest that increasingly concise instructions reflect shifts in \textit{referential} words.



\paragraph{More abstract referring expressions across repetitions}
A natural explanation for the shift in the words used is that Architects had learned to produce referring expressions at a higher level of abstraction, in particular ones that corresponded to entire towers rather than individual blocks.
To evaluate this possibility, the same human annotators additionally tagged each referring expression with the number of references to block-level and tower-level entities they contained.
Unsurprisingly, given that there were eight blocks in each scene and only two towers, we found that the number of references to blocks was greater overall than those made to towers ($b = -7.41$, $t(2344) = -20.98$, $p<0.001$), Fig.~\ref{fig:refexp}B). 
More importantly, we found that these proportions shifted across repetitions ($b = 1.35$, $t(2344) = 10.49$, $p<0.001$; interaction between repetition number and reference type). To measure this change in proportion, utterances in each trial were tagged as containing block-specific (e.g. ``horizontal blue block,'' ``vertical red block''), tower-specific (e.g. ``C shape,'' ``L shape''), or mixed expressions. Fig.~\ref{fig:refexp}C shows this change in proportion; reflecting both an increase in the number of tower-level references  and corresponding decrease in the number of block-level references.

\paragraph{Consistency and variability in referring expressions across dyads}
The overall increase in tokens resembling entire towers (``C'' and ``L'' shapes) in the final repetition suggests some degree of consistency between dyads, with respect to the tower-level abstractions that emerged.
To what extent did different dyads converge on the same set of labels for each tower, rather than settle on distinct, but internally consistent ways of referring to them?
To explore this question, we estimated how dissimilar the language used by different dyads was within each repetition, by computing the Jensen-Shannon divergence (JSD) between their word frequency distributions, aggregating language from all trials in a repetition block. 
We found that the mean pairwise JSD increased significantly between the first and final repetitions ($d=0.080$, 95\% CI:$[0.041 , 0.118]$, $p=0.004$), consistent with divergence between dyads.
We visualized these distances using a t-SNE embedding of word count vectors (Fig.~\ref{fig:refexp}C), revealing that this divergence might be attributed to the formation of distinct ``clusters'' (denoted by different colors shown with representative words; gray dots belong to degenerate clusters with $<4$ members).
Together, these findings suggest that even in this relatively simple task domain, human dyads manage to discover a diverse array of solutions for mapping tokens of natural language to components of each scene.


\begin{figure*}[t]
\begin{center}
\includegraphics[width=0.6\textwidth]{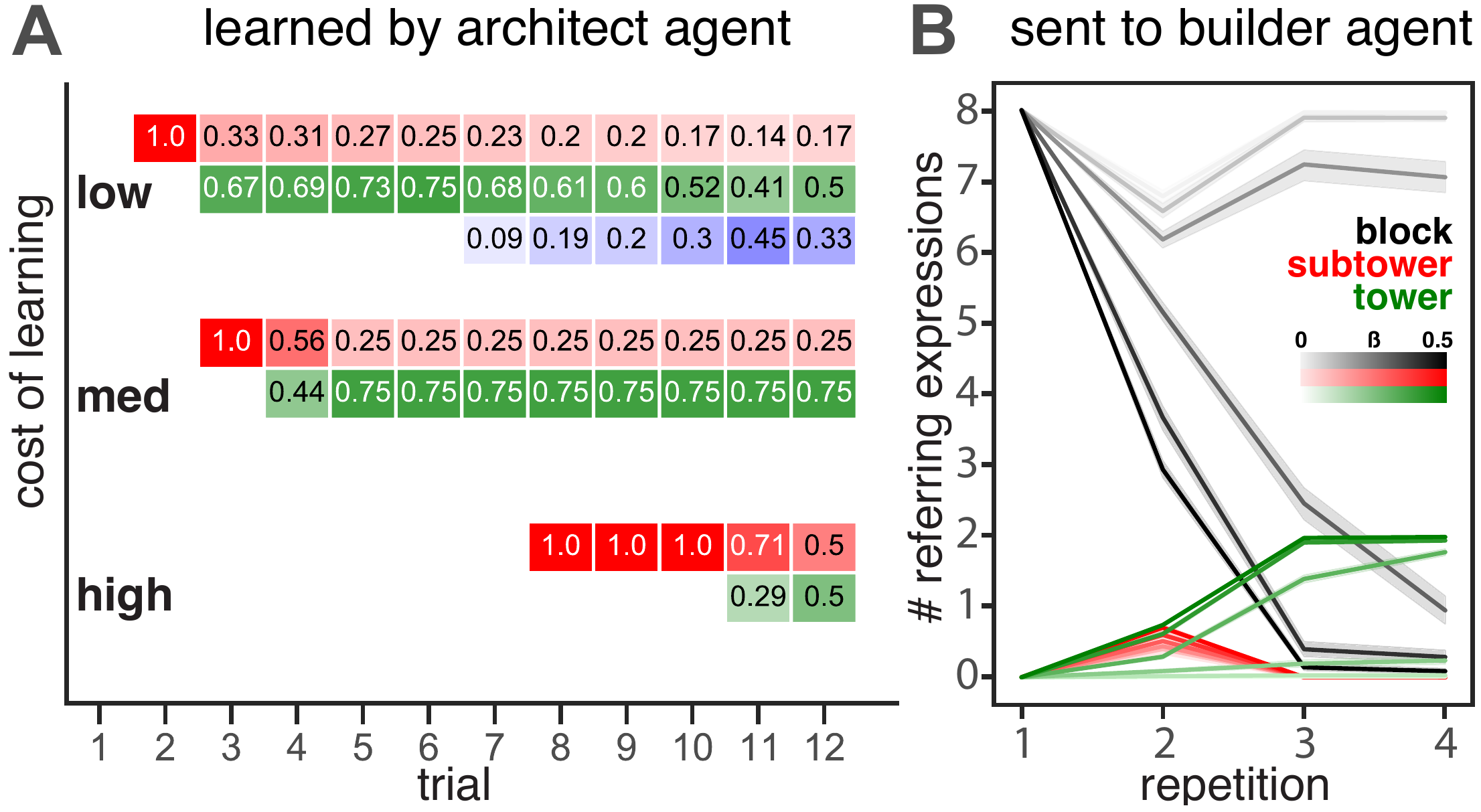}
\caption{(A) Trajectory of new procedural primitives added to the agent's library over the course of the task, shown for library size penalty $w=1.5$ (low), $w=3.2$ (medium), and $w=9.6$ (high). Each row represents the proportion of fragments at the \emph{sub-tower} level (red), \emph{tower} level (green), or \emph{scene} level (blue). Values in each cell represent the proportion of the agent's abstractions at that level. (B). The Architect model's production preferences over repetition blocks, shown for varying levels of cost-sensitivity parameter $\beta$, where $\beta=0.3$ best matches human data.}
\label{fig:model}
\end{center}
\end{figure*}

\section{Computational model}

In the previous section we found that Architects shift to more abstract tower-level referring expressions over successive interactions.
But why did participants generally introduce new words referring to entire \emph{towers} as units, as opposed to sub-towers or entire scenes?
Furthermore, given that initial reconstruction accuracy was already so high, why did participants decide to introduce new words at all?
We hypothesized that Architects' use of abstract referring expressions was constrained by the procedural abstractions available to each agent at a given time, as well as a rational trade-off between efficiency and informativity.
In other words, the Architect must (1) have an underlying representation of the procedure they intend to communicate, (2) maintain uncertainty about whether the Builder is likely to share that representation, and (3) prefer shorter message over longer messages, all else being equal.

We formalize this hypothesis in a computational model that integrates a Bayesian program learning algorithm  \cite{ellis2020dreamcoder} with a probabilistic model of communication under uncertainty and verify that these mechanisms give rise to the behaviors observed in our empirical data.
Repeated exposure to target towers increases the likelihood that chunked subroutines at the tower-level will be discovered by each agent.
This means that over the course of interaction, as the Architect becomes more confident that their abstracted referring expressions will be interpreted correctly, they increasingly prefer more efficient descriptions.

\paragraph{Procedural abstraction as program learning}

\newcommand{\mdl}{\mathsf{MDL}}
\newcommand{\size}{\mathsf{size}}

We begin by specifying how each agent's procedural knowledge is represented and modified over the course of learning in the task.
Following \citeA{ellis2020dreamcoder}, we assume that each agent maintains a library $\mathcal{L}$ of primitives that can be combined to generate simple block structures in a domain-specific language (DSL).
We assume the library is initialized with the following primitives: \texttt{h} (place a horizontal block), \texttt{v} (place a vertical block), \texttt{l} (move hand to the left), \texttt{r} (move hand to the right) and digits \texttt{1}$\sim$\texttt{9}. 
This DSL is small but fully expressive: any possible tower can be written by combining together these basic commands.

In the Bayesian program learning framework, the DSL is updated over time by expanding the library with new primitives.
As an agent progresses through multiple trials of tower scenes $\{T_{n}\}_{1}^{N}$, they may extract common subroutines that would allow them to re-represent the data more efficiently.
Formally, the model proposes a set of candidate sub-routine fragments $f$ after each trial and updates a posterior distribution over possible ways of extending the library (including $f = \emptyset$, which would maintain the current library):

\begin{equation}
P(\mathcal{L} \cup \{f\}|\{T_{n}\}_{1}^{N}) \propto {\underbrace{P(\mathcal{L} \cup \{f\})}_{\mathclap{\text{description-length prior}}}} \quad \times {\underbrace{\prod_{n=1}^{N}P(T_{n}|\mathcal{L} \cup \{f\})}_{\mathclap{\text{likelihood}}}}
\label{eq:posterior}
\end{equation}

This posterior distribution weighs two competing criteria for a good library: the likelihood and the prior.
The \emph{likelihood} in \ref{eq:posterior} captures the ability of an extended library efficiently to explain previous towers:
\begin{equation*}
   P(T_n | \mathcal{L}\cup \{f\}) = \mathsf{exp}(- \mdl(T_{n}\mid \mathcal{L} \cup \{f\}))
\end{equation*}
where $\mdl$ is a function evaluating the \emph{minimum description length}.
Intuitively, the MDL is the most compact version of $T_{n}$
that can possibly be written in the updated library $\mathcal{L} \cup \{f\}$.
This term is therefore maximized by sets of fragments $\{f\}$ that allow the existing data to be expressed most efficiently. 

The \emph{prior}, on the other hand, captures a preference for smaller libraries:
\begin{equation*}
   P(\mathcal{L} \cup \{f\}) = \mathsf{exp}(- w \cdot \size(\mathcal{L} \cup \{f\}))
\end{equation*}
where $\size(\mathcal{L} \cup \{f\})$ represents the number of primitives in the updated library. 
The strength of this preference is controlled by a parameter $w$.
We explore several values of $w$ in our simulations. 
Intuitively, when $w=0$, there is no penalty for having a larger library, so the library that best explains the observations would simply be the exhaustive set of scenes $T_n$ observations themselves.
As $w \rightarrow \infty$, any expansion of the library is considered too costly, preventing library learning entirely. 
These two objectives balance out in the posterior distribution (Eq. \ref{eq:posterior}) such that the fragments $f$ with the highest posterior probability are those that provide maximal compression of input tower programs while minimizing expansion of the library.

In practice, we selected the single highest posterior-probability set of fragments at each point in the task, conditioning on the previous trials (Fig.~\ref{fig:model}A).
The resulting DSL was supplied to both the Architect and Builder agent model as the set of primitives they are able to represent.
In other words, we assume that the Builder and Architect learn abstractions at the same rate throughout the experiment.
We further assume that when the Architect agent is presented with a scene, they are able to synthesize a set of 1 to 4 possible candidate programs for representing that scene in their current DSL. 
For example, the Architect agent may simultaneously recognize that a scene may be constructed by placing eight primitive blocks, \texttt{(h (l 1) v v (r 2) ...)}, or by combining two higher-level primitives \texttt{(chunk1 (r 2) chunk2)}. 


\paragraph{Communication as social reasoning}

In this section, we present a model of communication where each agent's DSL serves as a basis for grounding structured linguistic meanings.
We assume the Architect is a cooperative speaker agent who aims to produce utterances that will allow the Builder agent to re-produce the target tower. 
For simplicity, the Architect generates natural language instructions \emph{sequentially}, aiming to produce an utterance that convey each step $t_i$ of a full procedural sequence $T$ written in their current DSL.
Following recent probabilistic models of communication as social reasoning \cite<e.g.>{GoodmanFrank16_RSATiCS}, they choose an utterance proportional to its communicative utility, based on whether the Builder is expected to take the intended action after hearing the utterance: 
\begin{align}
P_S(u | t_i) & \propto \exp\{- \alpha \cdot U(u; t_i)\}\label{eq:rsa}
 \\
U(u; t_i)  & = \log P_L(t_i | u) \nonumber\\
P_L(t_i | u) & \propto \delta_{\llbracket u \rrbracket (t_i)}\nonumber
\end{align}
$\delta_{\llbracket u \rrbracket (t_i)}$ is the literal meaning function that the Builder agent is expected to use, evaluating to 1 when $u$ is true of the primitive $t_i$ in the agent's lexicon and 0 otherwise. 

The key behavioral phenomenon we aimed to explain with this model is the Architect's increasing preference for more abstract descriptions (i.e. Fig.~\ref{fig:refexp}B). 
We hypothesized that this behavior is a consequence of a rational trade-off between informativity and the cost of communication.
While Eq.~\ref{eq:rsa} gives the Architect's preferences for conveying each instruction of a fixed program $T$, we showed in the previous section that an Architect on later trials in fact has multiple ways of representing the raw scene $T^*$ available to them, using different primitives in their library.
We therefore extend our model to explicitly model the Architect's joint decision over which of these \emph{programs} $T^k$ to attempt to transmit in addition to what utterance they should use to transmit it:
\begin{align}
U(u, T^k; T^*) = (1-\beta) \cdot \sum_{i} \ln P_L(t_i^k | u) - \beta\cdot |T^k|
\end{align}
where $\beta$ is a parameter controlling the Architect agent's cost-sensitivity: when $\beta$ is high, the length of the required description dominates the Architect agent's decision-making; when it is low, the Architect's decisions are solely based on informativity to the Builder.

Finally, to account for the last condition of our hypothesis, that the Architect is sensitive to the risks of introducing novel descriptions, we must say what the meaning of a novel word should be in the Builder agent's lexicon: $\llbracket u \rrbracket$. 
Following recent models of convention and coordination \cite{hawkinsgeneralizing}, we assume that the Architect actually maintains uncertainty over the lexical mappings between words and primitives in the DSL $P(\llbracket u \rrbracket)$ and marginalizes over this distribution when evaluating their utility.
Some basic entries are deterministic, e.g. \texttt{\{h : ``place a horizontal block''\}}, but for learned abstractions (\texttt{chunk1}, \texttt{chunk2}), we assume a uniform distribution over an additional set of synthetic tokens (\texttt{``chunkA''}, \texttt{``chunkB''}) that can be emitted.
Over successive trials, the Architect agent can observe the Builder agent's actions (e.g. their placement of blocks) and update their beliefs about the lexicon \cite<see>[for additional details]{hawkinsgeneralizing}. 

\subsection{Simulation results}

\paragraph{The emergence of tower-level fragments}
Before presenting our Architect simulations, we first examine the trajectory of procedural abstractions that were acquired by the model over the 49 trial sequences presented to participants, while varying the penalty on library size, $w$ (Fig.~\ref{fig:model}A).
We manually categorized the resulting fragments based on their level of abstraction at the \emph{sub-tower} level (e.g. a routine producing a configuration of 2-3 blocks that co-occur within multiple towers), the \emph{tower} level (e.g. a routine generating four block placements that exactly reproduce one of the tower stimuli), or the \emph{scene} level (e.g. a routine generating eight block placements in the exact configuration that appeared on a trial).
First, we found that the statistical structure of the trial sequence did indeed allow our library learning algorithm to acquire full \emph{tower-level} primitives across a wide range of $w$, although higher (e.g. $w=9.6$) significantly delayed learning.
Surprisingly, the discovery of tower-level fragments was always preceded by sub-tower fragments.
For example, the pair of blocks forming the lower left of the 'L' and 'C' towers was frequently added, and many more such fragments were added at lower values of $w$.
There are several possible reason why these sub-tower abstractions were rare in our behavioral data, and additional work is required to determine whether Architects failed to represent them as perceptual configurations, or whether they simply suppressed the production of referring expressions for such structures. 

\vspace{1mm}
\subsubsection{Cost-sensitive Architects increasingly prefer abstract descriptions}
Next, we examine the results of a simple simulation exploring the dynamics of interaction between our Architect and Builder models.
We ran 2 iterations of each trial sequence, sampling an intended program and sequence of instructions from the Architect agent's distribution in Eq.~\ref{eq:rsa} and then sampling a set of resulting actions from the Builder agent's distribution conditioned on this utterance.
The agents updated their DSL and their beliefs about the lexicon after each trial. 
We found that Architects with strong cost-sensitivity (i.e. $\beta>0.5$) always used the most concise programs available to them -- by the third repetition nearly all block-level instructions were replaced by descriptions at higher levels of abstraction, even though these descriptions were more likely to result in Builder errors (Fig.~\ref{fig:model}B).
Meanwhile, in the absence of cost-sensitivity ($\beta=0$), Architects preferred a safer strategy, continuing to use longer but less ambiguous descriptions composed of block-level instructions even though more abstract representations were available to them (Fig.~\ref{fig:model}B, grey lines).
We found that intermediate values of $\beta$ roughly reproduced the qualitative Architect behavior observed in our behavioral data.



\section{Discussion}



Successful collaboration in many real-world tasks requires coordinating on shared abstractions and the language used to express them. 
This paper investigated how humans efficiently collaborate in a physical assembly task by developing shared abstractions for connecting language with object representations. 
We found that, across repeated interaction, dyads developed increasingly efficient communication by shifting language to more abstract referring expressions, without sacrificing communicative accuracy. 
We also implemented a computational model that integrates Bayesian program learning with a probabilistic model of communication to show how efficient abstract descriptions arise from a trade off between informativity and the cost of communication. 




Our approach extends a recent framework for studying convention formation \cite{HawkinsFrankGoodman17_ConventionFormation} to a task that requires performing complex procedures.
Communicating about these procedures is costly, making this task well suited to studying the emergence of abstractions under functional constraints such as informativity and efficiency.
By representing procedures as learned program fragments, our model provides a natural quantification of communicative efficiency-- i.e. program length, as well as an explicit mechanism for abstraction learning.
Together, this approach sheds light on few-shot and one-shot learning exhibited by humans as they use language to coordinate on new tasks.




In future work, we plan to further investigate the sources of consistency and variability in the communication protocols that emerge during collaboration and refine the program learning algorithm to fit our experimental results.
A currently unexploited source of consistency is in the choice of referring expressions used to refer to each tower (e.g.``C-shape'', ``upside-down U'').
An immediate next step will be to collect realistic priors for the expressions used to refer to entities in this stimulus set, in order to more precisely track uncertainty over the interpreted meaning of newly-formed conventions.
Another empirical observation our model does not capture is explicit reference to previous trials, which may be particularly relevant, as it appeared to frequently coincide with the emergence of expressions that refer to entire towers.

Furthermore, we made the simplifying assumption that both participants learned internal representations of procedural abstractions at the same rate.
While this may be the case in our highly structured building experiment, people collaborating on tasks in the real world are likely to discover useful abstractions at different times, due to differences in prior knowledge and from approaching the task from different perspectives.
While our model architecture posited a clean separation between the discovery of conceptual abstractions and their subsequent communication, people may actually leverage language to discover new abstractions, a possibility we are currently exploring by extending the library-learning component of our model with a SOTA Bayesian program learning algorithm that incorporates language. 

We have studied how the emergence of effective communication protocols over very sparse interaction allows humans to coordinate to solve physical assembly problems. Fruitful extensions could probe more complex domains, include artificial agents, or explore other algorithmic approaches (e.g., program synthesis, reinforcement learning, \textit{seq2seq}, etc.) to explain the computational mechanisms that enable effective coordination.
In the long term, such studies may shed light on the inductive biases that enable rapid coordination upon shared procedural abstractions during social interaction between intelligent, autonomous agents.

\section{Acknowledgments}

Thanks to the members of the Cognitive Tools Lab at UC San Diego for helpful discussion. C.H. is supported by a DoD NDSEG Fellowship. This work was supported by NSF CAREER Award \#2047191 to J.E.F.

\vspace{2em}
\fbox{\parbox[b][][c]{7.3cm}{\centering {All code and materials available at: \\
\href{https://github.com/cogtoolslab/compositional-abstractions}{\url{https://github.com/cogtoolslab/compositional-abstractions}}
}}}
\vspace{2em} \noindent

\bibliographystyle{apacite}

\setlength{\bibleftmargin}{.125in}
\setlength{\bibindent}{-\bibleftmargin}

\bibliography{references.bib}

\end{document}